\documentclass[runningheads]{llncs}

\usepackage{accv}

\usepackage{accvabbrv}

\usepackage{graphicx}
\graphicspath{{./}{./figures/}}
\usepackage{booktabs}
\usepackage{multirow}
\usepackage{multicol}
\usepackage{colortbl}

\definecolor{cvjepa}{RGB}{224,240,228}   %
\definecolor{cnovel}{RGB}{255,243,205}   %
\definecolor{cvmae}{RGB}{222,238,242}
\definecolor{cvl3}{RGB}{226,232,246}
\definecolor{cdino}{RGB}{238,228,244}
\definecolor{cintern}{RGB}{250,238,222}
\definecolor{cqwen}{RGB}{248,228,228}

\usepackage[accsupp]{axessibility}  %

\usepackage{hyperref}

\usepackage{orcidlink}

\begin{document}

\title{GTASA: Ground Truth Annotations for Spatiotemporal Analysis, \\ Evaluation and Training of Video Models}

\titlerunning{GTASA: Ground Truth Annotations for Spatiotemporal Analysis}

\author{Nicolae Cudlenco\inst{1,4}\orcidlink{0000-0001-6547-3659} \and
Mihai Masala\inst{2}\orcidlink{0000-0003-3496-9058} \and
Marius Leordeanu\inst{1,2,3}\orcidlink{0000-0001-8479-8758}}

\authorrunning{N.~Cudlenco et al.}

\institute{Institute of Mathematics of the Romanian Academy, Bucharest, Romania\\
\email{\{nicolae.cudlenco,leordeanu\}@gmail.com} \and
National University of Science and Technology Politehnica Bucharest, Romania\\
\email{mihaimasala@gmail.com} \and
Norwegian Research Center (NORCE), Bergen, Norway\\
\url{norceresearch.no} \and
B\"uchi Labortechnik AG, Flawil, Switzerland\\
\email{cudlenco.n@buchi.com}}

\maketitle

\begin{abstract}
Game engines hold what video models struggle to learn: a complete, explicit world state behind every frame. We turn one into a data
instrument. GEST-Engine, our production-grade open-source system, deterministically executes Graphs of Events in Space and Time (GESTs),
whether procedurally generated or derived from text, into videos of synchronized multi-actor scenarios, recording ground truth as it renders:
3D entity and camera state, pairwise spatial relations, event-to-frame mappings, instance segmentation, and long descriptions, at zero marginal
annotation cost. With it we release GTASA, a 938-video sample of what the system can generate at arbitrary scale, carrying,
to our knowledge, \textbf{the densest spatial-relation coverage of any video dataset}: a complete entity-pair relation graph
at every frame, $\sim$84$\times$ denser than the state of the art, frame-for-frame. We validate GTASA both qualitatively, through human evaluation of physical
validity and semantic alignment where frontier neural generators, given the same prompts, largely fail, and quantitatively,
with GTASA pretraining improving VideoLLaMA3 video captioning. Probing six frozen video encoders across 11 spatio-temporal tasks
enabled by GTASA's exact 3D ground truth, \textbf{a previously untestable inter-entity relational probe of frozen video features},
reveals that who-is-near-whom barely rises above chance for all of them. We release the engine, the corpus, and the benchmark, making this gap a measurable, trainable target.
\end{abstract}

\section{Introduction}
\label{sec:intro}

\begin{figure}[!t]
   \centering
   \includegraphics[width=\textwidth]{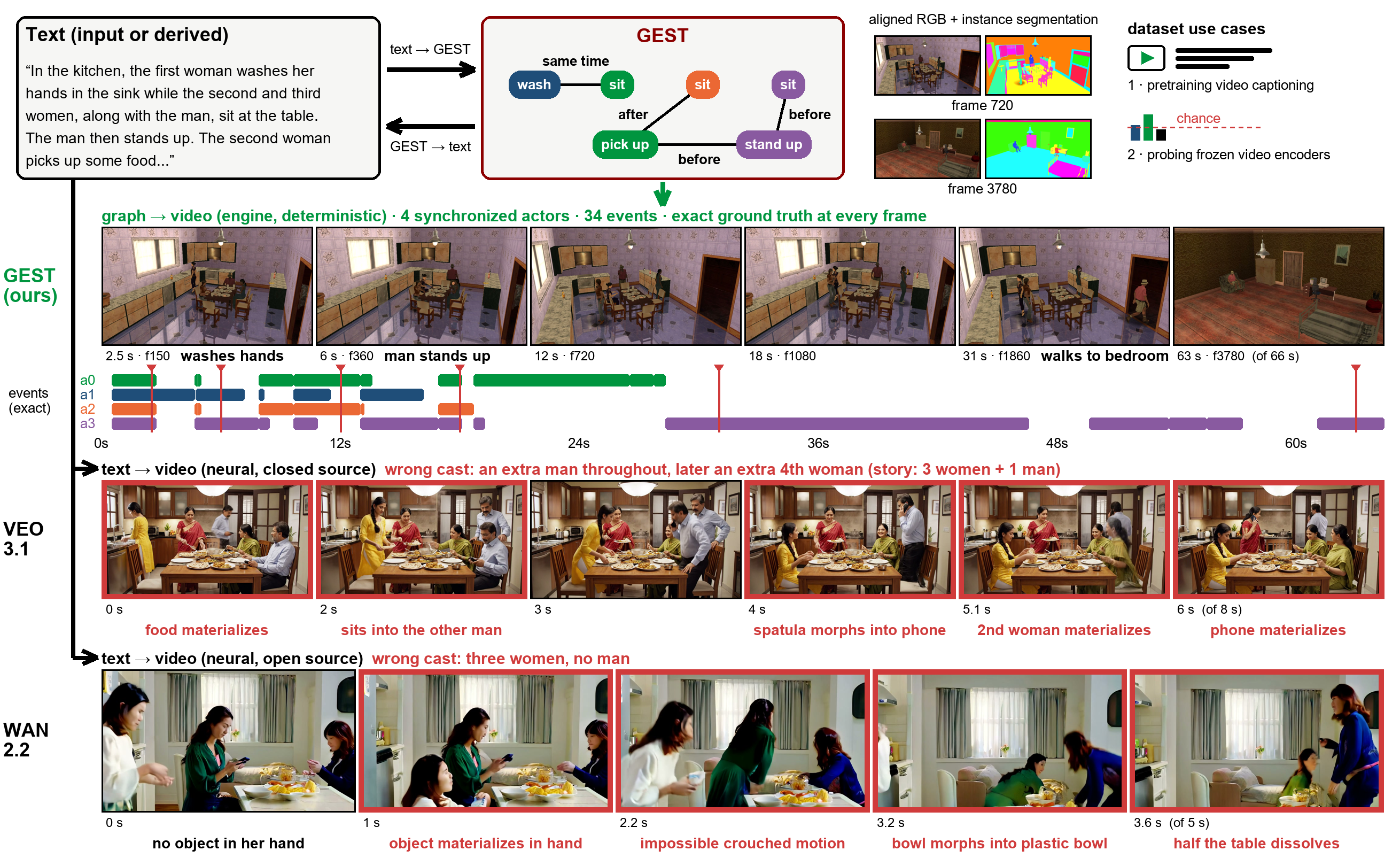}
   \caption{\textbf{GEST-Engine} turns graph or text specifications into multi-actor scenario videos with frame-synchronized spatial, temporal, visual, semantic, and textual ground truth from its explicit world model. The human-validated data enables VLM pretraining and spatio-temporal benchmarking. Here, from the GEST (top), it renders 66\,s with four synchronized actors and 34 events mapped exactly to frames (green). The same graph-derived text prompts VEO~3.1 and WAN~2.2, whose hand-verified outputs miscast the scenario and break object permanence and physics (red). Full videos are in the supplementary video (\url{https://youtu.be/ElYuoQgF17s}).}
   \label{fig:hero}
\end{figure}

Can we leverage the explicit world models of 3D game engines for video generation, video understanding, and benchmarking? Inside such an engine, every entity has an exact 3D position, every action a precise beginning and end, and the engine knows which entity each pixel belongs to. Videos, by contrast, are only a 2D capture: the spatial structure, entity identities, and occlusion relationships the engine maintains explicitly are lost in the projection. We turn one such engine into a data instrument: it executes synchronized multi-actor scenarios and records the world state as it renders, producing videos paired with ground truth that annotation pipelines recover only partially and at great cost (Fig.~\ref{fig:hero}).

Neural video generators, from open-source Wan \cite{wan2025wan} and LTX-Video \cite{hacohen2024ltx,hacohen2026ltx} to closed-source Sora \cite{brooks2024sora} and VEO 3 \cite{veo3modelcard,veo3announcement,veo3launch}, now produce photorealistic videos of cinematic quality. In parallel, neural world models, from foundational work \cite{ha2018worldmodels} to DreamerV3 \cite{hafner2023dreamerv3}, DIAMOND \cite{alonso2024diamond}, and Genie \cite{bruce2024genie,genie3}, learn increasingly powerful implicit representations of environment dynamics \cite{liu2025worldsurvey,lecun2022path}. Yet video understanding models still struggle with fine-grained temporal reasoning \cite{cai2024temporalbench,mangalam2023egoschema}, and video generators exhibit spontaneous appearance and disappearance of objects \cite{brooks2024sora} and physically implausible interactions \cite{wan2025wan}.

Existing video datasets reflect a persistent tradeoff between scale and annotation richness (Table~\ref{tab:dataset_comparison}). Large-scale datasets provide millions of clips but only short captions, ASR transcripts, or coarse temporal labels \cite{kay2017kinetics,miech2019howto100m,chen2026action100m}. Clip-level captioning corpora pair a single action with a single sentence: MSVD~\cite{chen2011collecting}, MSR-VTT~\cite{xu2016msr}, and VATEX~\cite{wang2019vatex} describe ten-second clips in 7--15 words, and ActivityNet-Captions~\cite{krishna2017dense} localizes on average 3.65 sentences of 13.5 words per video. Recent corpora reach 145--500 words per video~\cite{yang2024vript,chen2024sharegpt4video,ju2024miradata,chai2025auroracap}, but the text is generated by vision--language models from sparsely sampled frames, anchored to no event-level ground truth. Densely annotated datasets offer spatio-temporal scene graphs at limited scale, with annotation noise and considerable human cost: Action Genome~\cite{ji2020actiongenome} attaches person-object graphs to five sampled frames per action, PVSG~\cite{yang2023panoptic} adds panoptic masks at similar density, MovieGraphs~\cite{vicol2018moviegraphs} annotates characters and their interactions on movie clips, and PSG4D~\cite{yang20234d} extends scene graphs to 4D, including on GTA-V footage. These annotations are descriptive, extracted from existing video, rather than prescriptive, specified before generation, and none records the exact 3D world behind the pixels. Synthetic platforms can produce perfect annotations, and their data transfers: purely synthetic training data has achieved state of the art on real-world benchmarks for human pose~\cite{black2023bedlam}, and synthetic combined with real outperforms real alone for driving scenes~\cite{richter2016playing}. These platforms, however, have so far lacked synchronized multi-actor coordination~\cite{dosovitskiy2017carla,puig2018virtualhome,qiu2023virtualhome}.

\begin{table}[!t]
  \caption{Comparison with video datasets and synthetic platforms. Our clips are action segments of 938 scenario videos. Spatial Rels: total pairwise spatial
  relations. Words: average
  words per video description, where reported. Ctrl.: controllable through a cross-actor synchronized event configuration before generation. Semantics: type of structured annotation of who does what, to/with what, and where. Per annotated
  frame, GEST-Engine records \textbf{84$\times$} more pairwise spatial relations (\textbf{615}, automated geometry-based relations) than Action
  Genome (7.3, hand-labeled semantic predicates); our total counts unique, \textbf{frame-deduplicated} relations.}
  \label{tab:dataset_comparison}
  \centering
  \scriptsize
  \setlength{\tabcolsep}{1.5pt}
  \begin{tabular}{@{}lrrcccccc@{}}
    \toprule
    Dataset & Clips & Hours & \begin{tabular}[b]{@{}c@{}}Spatial\\Rels\end{tabular} & Semantics & \begin{tabular}[b]{@{}c@{}}Temporal\\Ann.\end{tabular} & Words & Interact. & Ctrl. \\
    \midrule
    VideoMix22M \cite{assran2025vjepa2} & 22M & 1M+ & -- & -- & -- & -- & no & no \\
    Kinetics \cite{kay2017kinetics} & 650K & 1,800 & -- & -- & clip-level & -- & no & no \\
    Action100M \cite{chen2026action100m} & 147M & 128K & -- & action text & approx. & -- & no & no \\
    MSVD \cite{chen2011collecting} & 1,970 & -- & -- & -- & clip-level & 7 & no & no \\
    MSR-VTT \cite{xu2016msr} & 10K & 41 & -- & -- & clip-level & 9 & no & no \\
    VATEX \cite{wang2019vatex} & 41.3K & 115 & -- & -- & clip-level & 15 & no & no \\
    ActivityNet-Cap. \cite{krishna2017dense} & 20K & 849 & -- & -- & localized & 49 & no & no \\
    LLM-captioned & 12K-- & -- & -- & -- & -- & 145-- & -- & no \\
    \quad \cite{yang2024vript,chen2024sharegpt4video,ju2024miradata,chai2025auroracap} & 330K & & & & & 500 & & \\
    Action Genome \cite{ji2020actiongenome} & 10K & 82 & 1.7M & relations & 5 fr/action & -- & yes & no \\
    PVSG \cite{yang2023panoptic} & 400 & -- & 150K & scene graph & annotated & -- & yes & no \\
     & & & frames & & & & & \\
    MovieGraphs \cite{vicol2018moviegraphs} & 7.6K & -- & graphs & social graph & timestamps & 35 & yes & no \\
    Videos-to-Paragraphs~\cite{bogolin2020hierarchical} & 510 & 4 & -- & SVO & SVO-level & 40 & yes & no \\
    CARLA \cite{dosovitskiy2017carla} & -- & -- & -- & -- & -- & -- & no & no \\
    BEDLAM \cite{black2023bedlam} & 30K & -- & -- & -- & -- & -- & no & no \\
    VirtualHome \cite{puig2018virtualhome} & 2.8K & -- & -- & program & action-level & 21 & no & no \\
    VH Act. Genome \cite{qiu2023virtualhome} & 5.8K & -- & graphs & scene graph & action-level & -- & no & no \\
    GEST \cite{masala2023gest} & 25 & -- & -- & event graph & sequential & -- & no & no \\
    WildWorld \cite{li2026large} & 3.4K & 574 & -- & action IDs & action-level & -- & no & no \\
    \midrule
    GEST-Engine (ours) & 28.2K & 14.3 & \textbf{892M} & \textbf{event graph} & 29.6K & \textbf{153} & \textbf{sync.} & \textbf{yes} \\
     & & & & & concurrent & & & \\
    \bottomrule
  \end{tabular}
\end{table}

These failures share a root: the world model behind the pixels is implicit and never persisted. Current models show signs of understanding, yet the longer the video, the higher the risk of losing coherence entirely. An explicit world model, a game engine or any other simulator, does not share this failure mode: every entity is simulated, and what to expect is known at every step, for any duration. Video encoders face the same question from the understanding side. Trained primarily on 2D video with reconstruction~\cite{wang2023videomae} or prediction~\cite{assran2025vjepa2} objectives, they aim to capture world properties such as physics, interaction, and cause and effect, yet it remains unclear whether they build an internal model with entities aligned in 3D space, or capture how inter-entity spatial relations evolve in time. If large-scale, exactly annotated data from an explicit world model were available, could we leverage it for world modeling, for training, and for answering such questions?

We built \textbf{GEST-Engine}, a production-grade system that closes this gap, and \textbf{GTASA}, the corpus it produced. GESTs, encoding actors, actions, objects, and their temporal and spatial constraints, were introduced to bridge vision and language, with video generation demonstrated on 25 sequentially ordered scenarios~\cite{masala2023gest}. GEST-Engine turns the representation into an instrument. In roughly 100K lines of code, most written before LLM coding assistants, developed over six years by a single researcher through open-source modding of a complete game world, it validates each specification, grounds abstract entities into concrete 3D assets, deterministically orchestrates actors under Allen-algebra temporal constraints~\cite{allen1983maintaining}, and records at every frame the entity and camera state, the complete pairwise spatial-relation graph, instance segmentation, and exact event boundaries, all at zero marginal annotation cost. GTASA comprises 938 such scenarios, a sample of what the system generates at arbitrary scale, each paired with a long description derived from its graph. The system itself, although mature and production ready, is a stepping stone rather than an endpoint: we open-source it so the community can extend it toward the full potential of the game world it builds on, or port its design to other engines.

\textbf{Q1\label{q1}: Is the generated data physically valid and semantically faithful?} We validate GTASA through human studies covering physical validity and text--video semantic match (Table~\ref{tab:human_evaluation_students}). As references we prompt frontier neural generators with the descriptions derived from the same graphs, text being the only specification they accept. Even when we additionally condition them on engine-rendered frames, the same failures persist, leaving open the challenge of integrating explicit world models with neural generators.

\textbf{Q2\label{q2}: Does GTASA transfer as training data?} We train VideoLLaMA3 2B and measure video captioning on real footage, where GTASA pretraining improves the model on all seven metrics (Tables~\ref{tab:captioning_results} and~\ref{tab:jury_results}).

\textbf{Q3\label{q3}: What do frozen video encoders know about space and time?} On GTASA's exact 3D ground truth we build a spatio-temporal probing benchmark of frozen video features, including inter-entity relational tasks that were previously untestable. Across eleven tasks and six encoders, we find that relations between entities are barely decodable from any of them, even by attentive probes, while video pretraining specifically buys motion understanding (Table~\ref{tab:spatial_probes}).

Our contributions are:
\textbf{(1)} We open-source GEST-Engine, a production system that compiles GESTs into videos of synchronized multi-actor scenarios with exact ground truth at every frame, extensible by design and documented in a full technical report~\cite{cudlenco2026gestengineeventgraphssynthetic}.
\textbf{(2)} We release GTASA: 938 procedurally generated videos of synchronized multi-actor scenarios, comprising 28.2K action clips, 892M pairwise spatial relations across 1.45M frames, and 29.6K exact event-to-frame temporal mappings, along with the generation pipeline for producing additional data at scale.
\textbf{(3)} We validate the data through human studies in which frontier generators, the only other systems accepting such synchronized multi-actor specifications, fail (\hyperref[q1]{Q1}), show captioning gains from GTASA pretraining (\hyperref[q2]{Q2}), and build a spatio-temporal benchmark of frozen video encoders with previously untestable inter-entity relational tasks (\hyperref[q3]{Q3}).

\section{System}
\label{sec:system}

\begin{figure}[!t]
   \centering
   \includegraphics[width=0.75\textwidth]{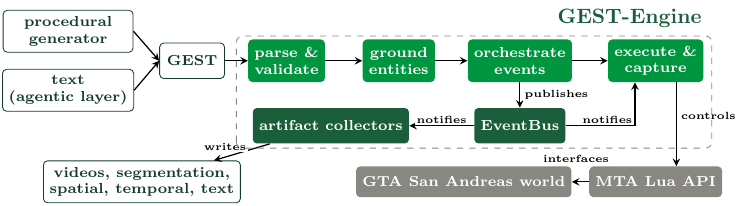}
   \caption{\textbf{GEST-Engine pipeline.} A GEST, procedurally generated or derived from text by an agentic layer, is parsed, validated, grounded, and orchestrated under temporal constraints. Execution in the MTA-scripted game world publishes events that frame-aligned collectors turn into the released artifacts.}
   \label{fig:architecture}
\end{figure}

Our pipeline proceeds in five stages: procedural graph generation, simulation execution, artifact collection, text generation, and corpus assembly (Fig.~\ref{fig:architecture}); full detail in the technical report~\cite{cudlenco2026gestengineeventgraphssynthetic}.

A procedural generator creates GEST specifications from the engine's capability registry: available episodes, points of interest, valid action chains, object types, and actor models. For each scenario it creates actors, builds per-actor action chains from the valid next-action sequences at each point of interest, synchronizes coordinated interactions and object exchanges through temporal constraints (before, after, same time), and moves random subsets of actors across regions. Because it only produces specifications the engine supports, every graph is executable by construction. Alternatively, an agentic layer derives GESTs from free text through tool calls validated by the same generator, so text-derived graphs inherit the same guarantee. This layer, makes the system, to our knowledge, the only one apart from neural video generators that takes free text as input and outputs video; program-driven simulators accept neither free text nor cross-actor temporal constraints.

Each GEST is passed to the simulation engine running within MTA (Multi Theft Auto)\footnote{https://multitheftauto.com} on GTA San Andreas, chosen when the project began six years ago because MTA offered mature open-source scripting control over a complete shipped game world. Reusing its assets, animations, and interiors made a system of this scope feasible for a team of one. The engine validates requirements against available environments, grounds abstract entities to concrete 3D objects, resolves temporal constraints through Floyd-Warshall transitive closure \cite{allen1983maintaining}, and dispatches actions to actor handlers with automatic camera tracking.

Corpus production runs unattended on a single 128GB workstation with 25 parallel virtual machines and a consumer GPU, at roughly 25 minutes per scenario and about 1TB per 1,000 videos. The architecture separates graph validation, entity matching, and temporal orchestration from concrete engine bindings, which supports porting to backends such as GTA V, Unreal, Unity, or any environment with programmatic control over actors and objects.

During simulation, a spatial relation collector queries the 3D position and rotation of every entity at each frame and computes relations between all entity pairs (Euclidean distance, relative direction, angular offsets), each entity retaining its scenario-level identifier from the input GEST. An event-frame collector records the exact frame at which each event begins and ends. The system produces instance segmentation (released with the corpus), depth, and pose.

We convert GEST representations into text using the two-stage approach proposed in~\cite{masala2025vision}: each graph is transformed into a proto-language, a concise machine-generated description, which an LLM\footnote{gpt-4o-2024-08-06} then refines into natural language without altering the core content (actors, events, order of actions), enriched with contextual details from the graph such as rooms and actor genders. The resulting descriptions serve as input for the captioning experiments (\hyperref[q2]{Q2}) and as prompts for the comparison videos (\hyperref[q1]{Q1}).

\section{The GTASA Dataset}
\label{sec:corpus}

In total we generate 938
(Video--GEST--Text) triplets.
Each GEST, mapped to the exact frames of its events, is an explicit state annotation at every frame: who does what, with or to whom or what, and where.
Table \ref{tab:corpus_stats} summarizes
GTASA. We also generated corresponding
videos with VEO 3.1 and WAN 2.2,
prompting them with the same
GEST-derived textual descriptions, the
only specification they accept. These
serve as references for the human
evaluation (\hyperref[q1]{Q1}) and as a
neural-generated training condition for
the captioning experiments
(\hyperref[q2]{Q2}).

At every frame, the collector logs the complete pairwise relation graph over all tracked entities, actors and objects alike: 78 relations per frame in the garden scene (13 entities), 528 inside the house (33), and 2,556 in the classroom (72), a corpus mean of 615. Frame for frame, this is roughly 84$\times$ the density of Action Genome~\cite{ji2020actiongenome} (7.3 per annotated frame), the closest existing comparator; per clip, 31.6K versus 170. The relations are exact geometric quantities (distance, direction, angular offsets) rather than curated semantic predicates, logged at every frame and stored change-deduplicated.
We publicly release the complete corpus (GEST, VEO, and WAN videos, spatial relations, temporal mappings, instance segmentation, descriptions), the generation pipeline, and the world editor with which we mapped the environments and objects, so the community can extend the game-world coverage and create new simulations.

\begin{table}[!tb]
  \caption{Corpus statistics}
  \label{tab:corpus_stats}
  \centering
  \scriptsize
  \setlength{\tabcolsep}{3pt}
  \begin{tabular}{@{}lr@{}}
    \toprule
    \textbf{Metric} & \textbf{Value} \\
    \midrule
    Scenarios (Video--GEST--Text triplets) & 938 \\
    Reference videos (VEO, WAN) & 938 each \\
    Total duration & 14.25 hours (median 48.7\,s per video) \\
    Action clips & 28.2K \\
    Annotated frames & 1.45M \\
    Events, with exact frame mappings & 29,579 \\
    Temporal relations & 12,457 \\
    Pairwise spatial relations & \textbf{892M} \\
    Actors per scenario & 2--6 (mean 3.43) \\
    Events per scenario & 7--65 (mean 29.4) \\
    Words per description & 153 (mean) \\
    \midrule
    Unique action types & 37 (social, manipulation, locomotion, exercise) \\
    Object types & 15 (furniture, devices, consumables, equipment) \\
    Environments & 10 episodes, 4 categories \\
    \midrule
    Available in the platform & 70+ environments, 733 objects, 2,500+ animations, 312 skins \\
    \bottomrule
  \end{tabular}
\end{table}

\section{Experiments}

\subsection{Data Validation (\hyperref[q1]{Q1})}
\label{sec:human_eval}

To address \hyperref[q1]{Q1}, we validate the generated data through an extensive human annotation process along two dimensions: video validity and text--video semantic alignment. All three systems (GEST-Engine, VEO 3.1, WAN 2.2) receive identical textual prompts derived from the same GEST graphs.

\textbf{Physical validity.}
Our focus is on real-world physical validity, with particular attention to objects and actors. We define an invalid video as ``one in which actors or objects appear, disappear, or transform into other objects without a logical, plausible explanation and in ways that would be impossible in reality''. We frame validity as binary classification.

\textbf{Semantic alignment.}
We assess how accurately the semantic content of the video corresponds to the input text, explicitly disregarding visual fidelity such as realism or texture quality and focusing solely on the depicted actions, the participants, and the sequence of events. We employ a five-point Likert scale from extremely low to perfect match. Note that these two tasks are independent: a video can be valid yet exhibit low semantic alignment with the input text, while an invalid video can have a strong semantic alignment with the input text.

\textbf{Annotators and assignment.}
All 16 annotators are volunteer graduate students. %
Videos were assigned in randomized scenario and system order, and every annotator received videos from all three systems under identical instructions, so any individual predisposition applies uniformly across conditions. The assignment maximized coverage, with a fraction of videos re-assigned to additional annotators to measure inter-annotator agreement. Guidelines and annotation interface are in Appendix~\ref{sec:annotation_details}.

\textbf{Results.}
Table~\ref{tab:human_evaluation_students} presents the results. Our method achieves a substantially higher validity rate than both generators. In our evaluation protocol, a video is invalid if any annotator marks it so, a conservative rule that can only lower validity rates. The validity rates for VEO 3.1 and WAN 2.2 are well below 20\%, which severely limits their practical applicability. The semantic score and average rank further show the stronger semantic alignment of our videos: 4.09 (out of 5) and rank 1.08 (1 is best).%

\begin{table}[!tb]
  \caption{Human studies. Left: validity ($\pm$~95\% confidence interval) and text--video semantic match. Right:
  reverse matching, where independent raters describe the videos and judges match
  them to the source prompts. Ours lead on all measures.}
  \label{tab:human_evaluation_students}
  \centering
  \scriptsize
  \setlength{\tabcolsep}{3pt}
  \begin{tabular}{@{}lccccc@{}}
    \toprule
    & \multicolumn{3}{c}{Direct study} & \multicolumn{2}{c}{Reverse matching} \\
    \cmidrule(lr){2-4} \cmidrule(lr){5-6}
    Category & Validity ($\uparrow$) & Semantic ($\uparrow$) & Rank ($\downarrow$) & Similarity ($\uparrow$) & Rank ($\downarrow$) \\
    \midrule
    GEST(G)-ours & \textbf{0.69}$\pm$.05 & \textbf{4.09} & \textbf{1.08} & \textbf{56.64} & \textbf{1.42}\\
    VEO (V) & 0.18$\pm$.05 & 2.50 & 2.04 & 42.94 & 1.86 \\
    WAN & 0.13$\pm$.04 & 1.75 & 2.88 & 23.44 & 2.72 \\
    \bottomrule
  \end{tabular}
\end{table}

\textbf{Validity is engineered.}
The 69\% validity rate is itself informative: even with an explicit world model, validity is engineered rather than inherited. The residual failures are execution artifacts, transient animation and rendering glitches during locomotion and object hand-overs, while the world state itself never loses an entity. The generators fail in the opposite and deeper way, breaking permanence and physics themselves. The gap to 100\% validity and to a perfect 5 measures the difficulty of faithful execution. The gap to 18\% and 13\% measures how much of that difficulty an explicit world model already solves.

\textbf{Inter-annotator agreement.}
Inter-annotator agreement was evaluated using both raw percentage agreement and Krippendorff's alpha~\cite{krippendorff2011computing}. For the validity task (binary classification), annotators achieved 75.8\% raw agreement ($\alpha$ = 0.488; nominal), in the range expected for a binary and moderately imbalanced task. For the semantic match task (rated on a 1--5 scale), exact agreement was 42.2\%. When allowing a tolerance of one scale point, agreement increased substantially to 85.4\%. Chance-corrected agreement was substantial ($\alpha$ = 0.636; ordinal), indicating that most disagreements were minor (i.e., differing by only one scale point) rather than reflecting fundamental semantic discrepancies.

\textbf{Reverse matching.}
As an independent validation, we perform a reverse matching experiment: raters describe each video without seeing the input prompt. These raters form a group fully independent from the first study, drawn from the same population, with no prior exposure to the videos or ground-truth data. Grouping the descriptions by source prompt yields, per prompt and annotator, one description for each of the three videos; the same judges as in Sec.~\ref{sec:captioning} then receive the original prompt as anchor and select the description that most closely matches it. The results (right side of Table~\ref{tab:human_evaluation_students}) closely align with the first study: our model performs best, followed by VEO 3.1 and WAN 2.2.

These results demonstrate that while neural generators produce visually realistic outputs, they frequently violate physical plausibility (82--87\% of the time) and fail to faithfully render the specified multi-actor scenarios. GEST-Engine's advantage is not visual realism but semantic reliability: the generated videos depict what was specified, with the correct actors, actions, and temporal ordering. The ranking is thus confirmed by three independent methods: the Likert study, reverse matching with a separate annotator pool and randomized assignment, and the VLMs-as-Jury evaluation of Sec.~\ref{sec:captioning}.

\textbf{Can frontier VLMs detect these failures?}
On the 94 videos with full annotator agreement (45 valid, 49 invalid), we prompt three frontier VLMs (GPT-5.2-Pro, Gemini-3.1-Pro, Claude-Opus-4.6) with the same instruction used in the human study to classify validity. All reach only 55--58\% accuracy against a 50\% baseline; a majority-vote jury reaches 58.5\%; full results in Appendix~\ref{sec:vlm_weak_points}. Objects appearing or disappearing without cause and actors morphing into different individuals are blind spots for current models. Moreover, when asked to describe such videos, the models ignore these inconsistencies and produce descriptions as if the content were coherent.

\begin{figure}[!t]
   \centering
   \includegraphics[width=\textwidth]{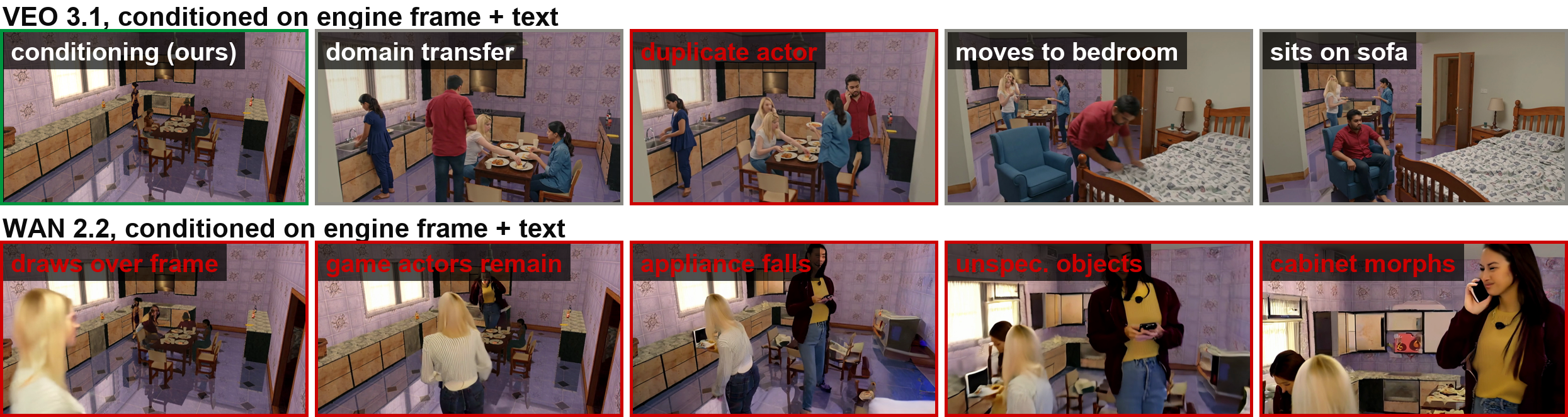}
   \caption{\textbf{Image conditioning does not transfer the world model.} Given an engine-rendered frame of the exact scene (green) and the graph-derived text, VEO~3.1 transfers the game environment into realistic footage and follows parts of the specification, yet materializes a duplicate of one of the actors (red). WAN~2.2 paints new actors over the frame without removing the game-rendered ones, then progressively breaks objects and geometry. Full videos are in the supplementary video.}
   \label{fig:conditioning}
\end{figure}

\textbf{Image conditioning.}
We additionally tested whether image conditioning can recover the failures: on 30 scenarios where both generators had failed under text-only prompts, VEO 3.1 and WAN 2.2 also received an engine-rendered frame of the exact scene, and three annotators evaluated all five conditions. Validity does not recover: WAN rates 3\% validity (semantic score 1.2) with conditioning versus 7\% (1.3) without, while VEO stays flat at 13\% validity with semantic match essentially unchanged (2.5 versus 2.3); the engine's corresponding scenarios rate 77\% (4.8). The conditioned outputs are nonetheless revealing (Fig.~\ref{fig:conditioning}): VEO sometimes performs a domain transfer of our game environment into realistic footage, re-rendering the room and following parts of the action sequence, yet still materializing unspecified actors, here a duplicate of an existing one; in other cases the output stays game-like or drifts toward cartoon styles. WAN instead paints new actors over the conditioning frame without removing the game-rendered ones, then progressively breaks objects and geometry. In none of our scenarios does conditioning improve validity or semantic match.

\textbf{Open challenge.}
Conditioning alone therefore does not transfer the world model, though it can transfer its appearance. We leave open the challenge of integrating the two families: an explicit world model provides exact state and guarantees, neural generators provide visual quality, and no current interface carries the former into the latter. Finer-grained conditioning, for example generating the scenario region by region, remains interesting future work.

\subsection{Video Captioning (\hyperref[q2]{Q2})}
\label{sec:captioning}

To address \hyperref[q2]{Q2}, we evaluate whether GEST-Engine data can serve as a training resource for video understanding models. We fine-tune VideoLLaMA3 2B \cite{zhang2025videollama}, an open-source vision--language model coupling a SigLIP vision tower with DiffFP temporal compression, under three conditions: (a) GEST then Real (G): Stage 1 fine-tuning on our GEST synthetic data followed by Stage 2 fine-tuning on real video, (b) VEO then Real (V): Stage 1 fine-tuning on VEO-generated data followed by Stage 2 fine-tuning on real video, and (c) Real-only (R): fine-tuning on real data only. We denote the human reference as H.

\textbf{Data.}
For Stage 1, we use the 938 GEST videos and the corresponding VEO 3.1 videos generated from identical textual specifications. As target real data, we use Videos-to-Paragraphs \cite{bogolin2020hierarchical}: real multi-actor sequences with dense two-level annotations, a structure that clip-caption corpora such as MSR-VTT and VATEX do not offer (see the dataset structure comparison in Appendix~\ref{sec:structure_comparison}). We use the official splits.

\textbf{Training.}
We use QLoRA \cite{dettmers2023qlora} (4-bit NF4, LoRA rank 16, alpha 32, dropout 0.05, applied to all linear layers) for all training due to hardware constraints (RTX 3090, 24GB). Stage 1 (fine-tuning on synthetic data) runs 5 epochs at a learning rate of 2e-4, while for Stage 2 (real data) the learning rate is halved and the number of epochs doubled; R trains directly on the real data for the same number of epochs at the default rate.%

\begin{table}[!tb]
  \caption{Video captioning metrics on Videos-to-Paragraphs for VideoLLaMA3 2B, mean$\pm$std over multiple random seeds. Best result in bold.}
  \label{tab:captioning_results}
  \centering
  \scriptsize
  \setlength{\tabcolsep}{3pt}
  \begin{tabular}{@{}lccccccc@{}}
    \toprule
    & BLEU-4 & METEOR & CIDEr & ROUGE-L & SPICE & BERTScore & BLEURT \\
    \midrule
    GEST(G)-Ours & \textbf{13.9$\pm$.5} & \textbf{35.8$\pm$1.1} & \textbf{29.2$\pm$3.6} & \textbf{37.2$\pm$.5} & \textbf{30.4$\pm$1.2} & \textbf{90.8$\pm$.1} & \textbf{54.7$\pm$.4} \\
    Real-only (R) & 12.2$\pm$.9 & 33.2$\pm$.9 & 25.5$\pm$5.9 & 35.3$\pm$.8 & 26.6$\pm$1.2 & 90.7$\pm$.1 & 53.3$\pm$.3 \\
    VEO (V) & 12.9$\pm$1.1 & 34.2$\pm$1.7 & 26.9$\pm$4.1 & 35.9$\pm$1.0 & 28.8$\pm$1.9 & 90.6$\pm$.2 & 53.9$\pm$.8 \\
    \bottomrule
  \end{tabular}
\end{table}

\textbf{Event-aware frame sampling.}
Leveraging GEST's ground-truth event-to-frame mappings, we implement a hybrid sampling strategy that selects mid-event frames for each action and fills the remaining slots (up to 64 frames) with uniformly sampled frames at 1 fps. We adopt hybrid sampling for all GEST conditions. Videos without event maps (real, VEO) are sampled at 1 fps.

\begin{table}[!b]
  \caption{Left: average ranks (lower is better), mean$\pm$std over 4 seeds; \textit{H}, the human reference, is rank 1 by construction in the metrics ranking (Jury is in agreement with humans); Right: head-to-head win rates (\%): Metrics = share of seed--metric comparisons won; Jury = per-clip majority vote of the three judges pooled across seeds. Conditions as in Table~\ref{tab:captioning_results}. Bold marks the winner.}
  \label{tab:jury_results}\label{tab:jury_winrates}
  \centering
  \scriptsize
  \setlength{\tabcolsep}{1pt}
  \begin{tabular}{@{}lcccc@{}}
    \toprule
    Ranking & Human & GEST & Real-only & VEO \\
    &(H) & (G)-Ours & (R) & (V) \\
    \midrule
    GPT-5.2 Pro & \textit{2.11$\pm$.1} & 2.63$\pm$.1 & \textbf{2.57$\pm$.2} & 2.69$\pm$.3 \\
    Gemini 3.1 & \textit{1.93$\pm$.2} & \textbf{2.58$\pm$.1} & 2.64$\pm$.2 & 2.84$\pm$.0 \\
    Claude 4.6 & \textit{2.25$\pm$.1} & \textbf{2.50$\pm$.1} & 2.63$\pm$.2 & 2.62$\pm$.3 \\
    \midrule
    VLMs as Jury & \textit{2.10$\pm$.1} & \textbf{2.57$\pm$.1} & 2.62$\pm$.1 & 2.72$\pm$.2 \\
    Metrics & \textit{1.00} & \textbf{2.00} & 3.86 & 3.14 \\
    \midrule
    Avg Rank & \textit{1.82} & \textbf{2.43} & 2.93 & 2.82 \\
    \bottomrule
  \end{tabular}
  \hspace{14pt}
  \begin{tabular}{@{}lcc@{}}
    \toprule
    Win rate & Metrics & Jury \\
    \midrule
    GEST vs VEO & \textbf{82}/18 & \textbf{53}/47\\
    GEST vs Real & \textbf{89}/11 & \textbf{51}/49 \\
    Real vs VEO & 39/\textbf{61} & \textbf{57}/43 \\
    \bottomrule
  \end{tabular}
\end{table}

\textbf{Results.}
Table \ref{tab:captioning_results} presents captioning metrics across all conditions, as mean and standard deviation over multiple random seeds. For comparability with the VLMs as Jury evaluation, we include humans (H) as rank 1 when computing average metric ranks. We evaluate using standard text similarity metrics~\cite{papineni2002bleu,banerjee2005meteor,vedantam2015cider,anderson2016spice,zhang2019bertscore,sellam2020bleurt}. Our GEST (G) achieves the best mean score on all seven metrics. The wins are consistent rather than borderline: counting each seed--metric pair as one comparison (Table~\ref{tab:jury_winrates}), GEST wins 89\% of the comparisons against Real-only and 82\% against VEO, where tied conditions would split near half.

\textbf{VLMs as Jury evaluation.}
Standard captioning metrics at small differences may not correlate well with human judgment. Following recent practice of using VLMs as evaluators \cite{zheng2023judging}, we assemble a jury of three frontier models (GPT-5.2 Pro, Gemini 3.1 Pro, Claude Opus 4.6) that independently rank the four captions (G, V, R, and H) for each test video. Table \ref{tab:jury_results} reports the per-judge average ranks, the combined rankings, and the head-to-head win rates on both bases. The jury prefers humans over every machine condition (62--63\% win rates), confirming that the evaluation is calibrated to human preference.

Among machine conditions, GEST ranks first on the average VLMs Jury rank and for two of the three judges, and wins both head-to-heads by per-clip majority vote, narrowly but in the same direction as the metrics (51\% to 49\% against Real-only, 53\% to 47\% against VEO). The jury thus serves as calibration: it puts Humans clearly first and GEST narrowly ahead, suggesting that the metric gains are not artifacts of reference wording. Prompts are in Appendix~\ref{sec:llm_jury}.

\textbf{Qualitative analysis.}
Beyond aggregate scores, we hand-verify a subset of caption sets against the test videos, with additional cases provided in Appendix~\ref{sec:qual_cases}. Our GEST-trained captions are visibly richer: in the verified case of Fig.~\ref{fig:qual_gexample2}, GEST recovers the full multi-actor sequence, captures a detail the human reference misses, and is ranked above that reference by the jury, while VEO and Real-only miss much of the story.%

These results suggest a benefit that is consistent across seeds and metrics and directionally confirmed by the VLMs as Jury, despite the visual domain gap: a system whose rendering is far from the state of the art produces more effective training pairs than the photorealistic output of a frontier generator. The value lies in the data's structure, not its pixels.

\subsection{Spatio-Temporal Probing of Video Encoders (\hyperref[q3]{Q3})}
\label{sec:probing}

\begin{figure}[!tb]
   \centering
   \includegraphics[width=\textwidth]{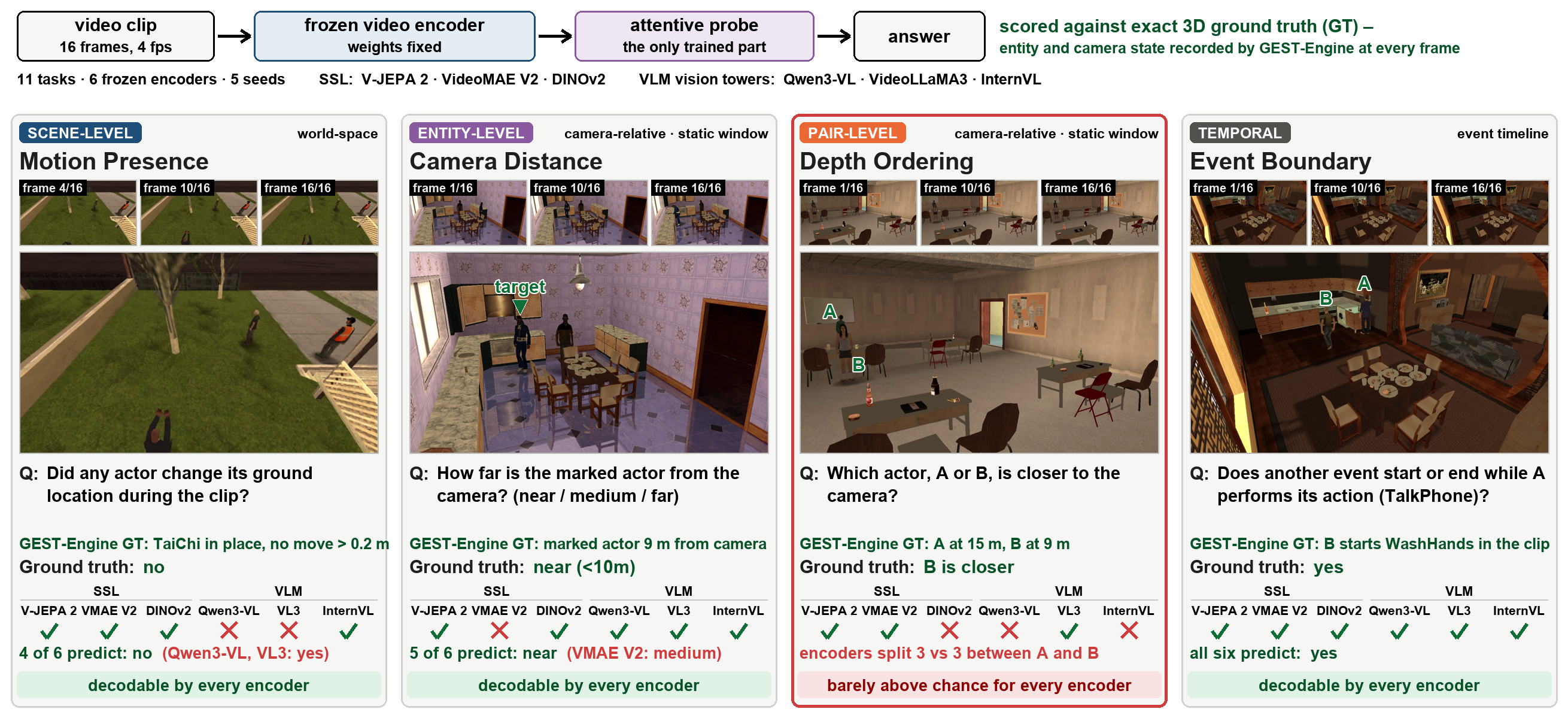}
   \caption{\textbf{Spatio-temporal probing of frozen video encoders.} A video clip passes through a frozen encoder, and an attentive probe, the only trained part, answers spatio-temporal questions scored against the engine's exact 3D ground truth(GT): eleven tasks in four categories, six encoders, five seeds. One example per category: scene-level and entity-level properties and event timing are decodable from every encoder, while pair-level relational questions (red) remain barely above chance for all six.}
   \label{fig:probes}
\end{figure}

To address \hyperref[q3]{Q3}, we probe what spatial and temporal information is encoded in frozen video encoder representations, leveraging GEST's exact 3D ground truth (Fig.~\ref{fig:probes}). We compare three self-supervised encoders, V-JEPA 2 \cite{assran2025vjepa2} (joint-embedding prediction), VideoMAE V2 \cite{wang2023videomae} (masked autoencoding), and DINOv2 \cite{oquab2023dinov2} (image-only self-supervision, included as a video-versus-image control), against the vision towers of three vision--language models: Qwen3-VL \cite{bai2025qwen3}, VideoLLaMA3 \cite{zhang2025videollama}, and InternVL2.5 \cite{chen2024expanding}. All six are ViT-based, but their training signals differ fundamentally, enabling a controlled comparison of how the training signal shapes spatial representation.

\textbf{Data.}
Each clip corresponds to a single GEST event. We sample 16 frames at 4 fps from each event segment, discarding events shorter than 4 seconds and excluding movement-only actions, for a total of 3,318 clips. The split is performed at the scenario level (70/15/15\% train/val/test), stratified by scene type, so no clip from the same scenario appears in multiple splits.

\textbf{Tasks.}
We define 11 probe tasks organized into four categories as follows:
\textbf{Scene-level Tasks} (clip classification): Actor Count (4-class: 2/3/4/5+ actors) and Motion Presence (binary: did any actor move).
\textbf{Entity-level Tasks} (one marked actor): Entity Presence (binary: is the entity visible), Camera Distance (3-class: near/medium/far), and Angle Change (binary: does the actor turn or keep its facing).
\textbf{Pair-level Tasks}: Depth Order (binary: which of two actors is closer to the camera), Pairwise Direction (multi-label over the four body-frame regions of the moving actor: front/back/left/right), Pairwise Distance (3-class: close/medium/far), Approach/Recede (3-class: toward/away/static relative to the camera), and Relative Motion (3-class: converging/diverging/static in world space).
\textbf{Temporal Tasks}: Event Boundary (binary: another event starts or ends in the clip).

Camera-relative tasks are re-sampled inside each clip's longest camera-static sub-window, so labels are actor-driven rather than artifacts of camera motion. Example clips are in the supplementary video.
Prior spatial probing evaluates dense per-pixel properties (depth, normals, correspondences) \cite{el2024probing,huang2025much}, rather than discrete entity-to-entity reasoning, and temporal benchmarks \cite{cai2024temporalbench,mangalam2023egoschema} measure reasoning failures on real video but cannot explain them, as no exact 3D state exists to probe against. Our benchmark sits at the intersection: frozen-feature probing of inter-entity relational geometry against exact per-frame 3D ground truth in dynamic multi-actor scenes.

\textbf{Probe architecture.}
Following the frozen evaluation protocol of V-JEPA 2 \cite{assran2025vjepa2}, we use an attentive probe inspired by theirs: a learnable query token attends to frozen encoder features through 4 cross-attention layers,
followed by a linear classification head. For entity-specific and pairwise tasks, we prepend learned entity-type embeddings to the query. All encoders are frozen; only probe parameters are trained. We use AdamW with cosine schedule, 5-epoch warmup, early stopping on validation balanced accuracy (BAcc), and report test BAcc.

\textbf{Results.}
Table \ref{tab:spatial_probes} presents balanced accuracy across all 11 tasks for the six encoders, as mean and standard deviation over five random seeds.

\begin{table}[!tb]
  \caption{Balanced accuracy (\%) on the spatio-temporal probing benchmark, mean$\pm$std over 5 seeds. Rnd = chance level. Best per task in bold. \colorbox{cnovel}{Highlighted} task names mark the previously untestable inter-entity relational tasks. Columns alternate the two families by within-family rank: V-JEPA 2 leads overall and VideoLLaMA3 (VL3) is the best VLM vision tower. SSL encoders: V-JEPA 2, VideoMAE V2 (video), DINOv2 (image); VLM vision towers: VL3, InternVL2.5, Qwen3-VL.}
  \label{tab:spatial_probes}
  \centering
  \scriptsize
  \setlength{\tabcolsep}{2.5pt}
  \begin{tabular}{@{}llc>{\columncolor{cvjepa}}c>{\columncolor{cvl3}}c>{\columncolor{cvmae}}c>{\columncolor{cintern}}c>{\columncolor{cdino}}c>{\columncolor{cqwen}}c@{}}
    \toprule
    Category & Task & Rnd & V-JEPA 2 & VL3 & VMAE V2 & InternVL & DINOv2 & Qwen3 \\
    \midrule
    Scene & Actor Count & 25.0 & 53.4$\pm$4.3 & \textbf{71.0$\pm$2.9} & 43.5$\pm$3.0 & 65.9$\pm$2.5 & 54.3$\pm$3.2 & 45.2$\pm$4.9 \\
    & Motion Pres. & 50.0 & \textbf{88.7$\pm$0.9} & 73.3$\pm$1.8 & 87.9$\pm$0.8 & 71.6$\pm$1.8 & 73.3$\pm$1.5 & 60.5$\pm$2.8 \\
    \midrule
    Entity & Entity Pres. & 50.0 & 85.1$\pm$2.6 & 82.9$\pm$1.6 & \textbf{86.0$\pm$1.6} & 85.3$\pm$1.2 & 85.4$\pm$1.0 & 83.6$\pm$1.6 \\
    & Camera Dist. & 33.3 & 75.0$\pm$2.4 & 77.8$\pm$1.5 & 76.5$\pm$2.6 & 78.0$\pm$0.8 & \textbf{79.4$\pm$1.3} & 77.3$\pm$1.9 \\
    & Angle Change & 50.0 & 81.3$\pm$1.4 & 62.8$\pm$7.7 & \textbf{81.4$\pm$3.9} & 55.4$\pm$2.4 & 58.6$\pm$4.5 & 53.7$\pm$6.4 \\
    \midrule
    Pair & \cellcolor{cnovel}Depth Order & 50.0 & 54.0$\pm$1.9 & \textbf{54.4$\pm$2.1} & 53.2$\pm$2.9 & 51.5$\pm$2.0 & 54.1$\pm$2.7 & 53.0$\pm$3.3 \\
    & \cellcolor{cnovel}Pair Direct. & 50.0 & 53.3$\pm$1.5 & \textbf{54.7$\pm$1.4} & 53.5$\pm$1.7 & 53.7$\pm$2.1 & 54.1$\pm$1.3 & 53.2$\pm$1.3 \\
    & \cellcolor{cnovel}Pair Dist. & 33.3 & 45.7$\pm$1.9 & \textbf{48.6$\pm$1.8} & 44.4$\pm$2.8 & 47.8$\pm$1.0 & 47.5$\pm$3.1 & 44.7$\pm$3.6 \\
    & Appr./Recede & 33.3 & \textbf{61.4$\pm$3.2} & 52.1$\pm$2.7 & 56.0$\pm$4.2 & 51.0$\pm$2.6 & 52.9$\pm$2.9 & 51.6$\pm$3.6 \\
    & Rel. Motion & 33.3 & \textbf{46.5$\pm$8.8} & 35.7$\pm$3.4 & 36.7$\pm$3.0 & 33.0$\pm$3.2 & 37.4$\pm$5.1 & 31.1$\pm$5.6 \\
    \midrule
    Temporal & Event Bound. & 50.0 & \textbf{78.7$\pm$1.2} & 76.3$\pm$2.1 & 77.1$\pm$1.0 & 75.1$\pm$1.4 & 74.6$\pm$2.5 & 73.2$\pm$1.5 \\
    \midrule
    \multicolumn{3}{@{}l}{\textit{Average}} & \textbf{65.7} & 62.7 & 63.3 & 60.7 & 61.1 & 57.0 \\
    \bottomrule
  \end{tabular}
\end{table}

\textit{Inter-entity relational geometry is barely decodable from any frozen encoder.} Depth Ordering (51--54\%) and Pairwise Direction (53--55\%) remain slightly above chance (50\%) for every encoder, video-SSL and VLM alike. Pairwise Distance reaches 44--49\% over a 33.3\% chance, yet stays roughly 30 points behind Camera Distance (75--79\%), the single-entity task with the same three classes. The remaining pair-level tasks, Approach/Recede and Relative Motion, probe motion rather than inter-entity relations. Who is in front of, near, or to the left of whom is barely decodable from these representations, even by an attentive probe, regardless of pretraining.

\textit{Scene, entity, and temporal properties are decodable.} All six encoders clear chance on Motion Presence (up to 88.7\%), Entity Presence (83--86\%), Camera Distance (75--79\%), Event Boundary (73--79\%), and Actor Count. Frozen features encode what is present, how many, how far, and whether something changed, but little of the geometry between entities. Counting is a known failure mode of language-supervised encoders \cite{paiss2023teaching}; VideoLLaMA3's SigLIP tower with temporal compression is the exception here (71\% on Actor Count).

\textit{Video pretraining, specifically, buys motion understanding.} V-JEPA 2 and VideoMAE V2 reach 88\% on Motion Presence and 81\% on Angle Change, while the VLM towers sit between 54\% and 74\%. Critically, DINOv2, the image-only self-supervised control, falls to VLM level on both (73.3\% on Motion Presence), isolating the advantage to video pretraining rather than self-supervision in general. VideoLLaMA3, the captioning model of Sec.~\ref{sec:captioning}, is also the strongest of the three VLM vision towers.

\section{Conclusion}
\label{sec:conclusion}

We open-source GEST-Engine, the capstone of a six-year engineering effort (presented in detail in the full technical report~\cite{cudlenco2026gestengineeventgraphssynthetic}), and GTASA, its $\sim$1TB sample corpus; three conclusions follow.

\textbf{First}, as training data for video captioning, GEST videos, at their current visual realism, outperform both real footage alone and the photorealistic output of VEO 3.1, the commercial state of the art, on all seven metrics and directionally under a VLM jury (Tables~\ref{tab:captioning_results}, \ref{tab:jury_winrates}); on the same specified scenarios, human raters prefer GEST videos for physical validity and semantic match by wide margins (Table~\ref{tab:human_evaluation_students}), exposing the generators' struggle: actors duplicating, objects vanishing.

\textbf{Second}, complete access to the world state (scenes, actors, events, 3D geometry, spatial relations, segmentation, depth, pose, the full story) makes the engine a ground-truth platform: exact video--text pairs for training and per-frame spatio-temporal supervision for probing what video encoders know (Table~\ref{tab:spatial_probes}).

\textbf{Third}, generation is fully controlled: the GEST comes procedurally generated, hand-crafted, or agentically derived from free text, and the open-sourced engine executes every event as specified, at story length.

The challenge we pose is to leverage the explicit world model to move the neural one. \textbf{This work contributes the explicit one.}

\textbf{Limitations and future work.} We continue working on stability fixes for the residual execution artifacts and towards better visuals, either through image realism enhancement on top of the current engine or through the adapter-pattern port to newer platforms, including GTA V. In parallel, we will grow the stories in length, complexity, and vocabulary of actions, objects, and environments.

\subsubsection*{Acknowledgements.}
The authors would like to thank Cristina Lazar for her valuable involvement in the human annotation process.
This work is supported in part by B\"uchi Labortechnik AG and by the projects
``Romanian Hub for Artificial Intelligence --- HRIA'', Smart Growth,
Digitization and Financial Instruments Program, 2021--2027 (MySMIS no.~334906)
and ``European Lighthouse of AI for Sustainability --- ELIAS'', Horizon Europe program (Grant No.~101120237).

\clearpage
\bibliographystyle{splncs04}
\bibliography{main}

\clearpage
\appendix

\section{Dataset Structure Comparison}
\label{sec:structure_comparison}

\begin{figure}[!tb]
   \centering
   \includegraphics[width=\textwidth]{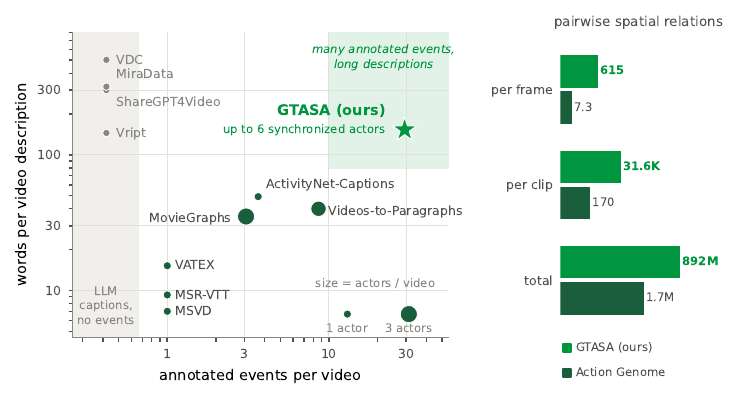}
   \caption{Per-video structure of captioning datasets: annotated events per video versus words per description (log--log). GTASA and Videos-to-Paragraphs occupy the multi-event, long-description region; clip-caption corpora cluster at single events and short sentences; corpora without event-level annotation are shown in the side band. Inset: per-annotated-frame spatial-relation density (Action Genome 7.3 vs.\ GTASA 615).}
   \label{fig:gap_scatter}
\end{figure}

The captioning experiments of the main paper (Sec.~4.2) use Videos-to-Paragraphs as the real target data. Fig.~\ref{fig:gap_scatter} situates this choice among captioning corpora by per-video structure. Clip-level corpora such as MSVD, MSR-VTT, and VATEX pair a single short clip, typically one actor performing one action, with a single sentence of 7--15 words; ActivityNet-Captions localizes on average 3.65 sentences of 13.5 words per video. Videos-to-Paragraphs is the closest real corpus to GTASA in structure: real multi-actor sequences carrying two-level annotations, SVO-level sentences aligned to action segments and paragraph-level stories of about 40 words. GTASA occupies the same multi-event, long-description region at larger scale: 2--6 actors (mean 3.43), 7--65 annotated events per video (mean 29.4), and 153-word descriptions. This structural match, rather than visual similarity, is why Videos-to-Paragraphs is the transfer target: the training signal of interest is multi-actor, multi-event description, which clip-caption corpora do not exercise.

\section{Annotation Details}
\label{sec:annotation_details}

In this section, we provide additional details regarding the annotation process. All annotators were independent participants recruited from the university student population; none of them is an author, a collaborator on this project, or a member of our laboratories, and none had prior exposure to the videos or the ground-truth data. Videos were presented anonymized and in randomized order, so annotators could not tell which system produced a given clip. The reverse-matching study of Sec.~4.1 of the main paper used a second, disjoint group recruited from the same population. It is important to note that at the beginning of the process we provided a live demo for all human annotators. During this session, we presented the task at hand, explained the annotation guidelines and introduced the live interface. The demonstration was designed to ensure that annotators clearly understood the task requirements and provided them with the opportunity to ask question before starting the annotation process.

\subsection{Interface}

\begin{figure}[tb]
   \centering
   \includegraphics[width=\linewidth]{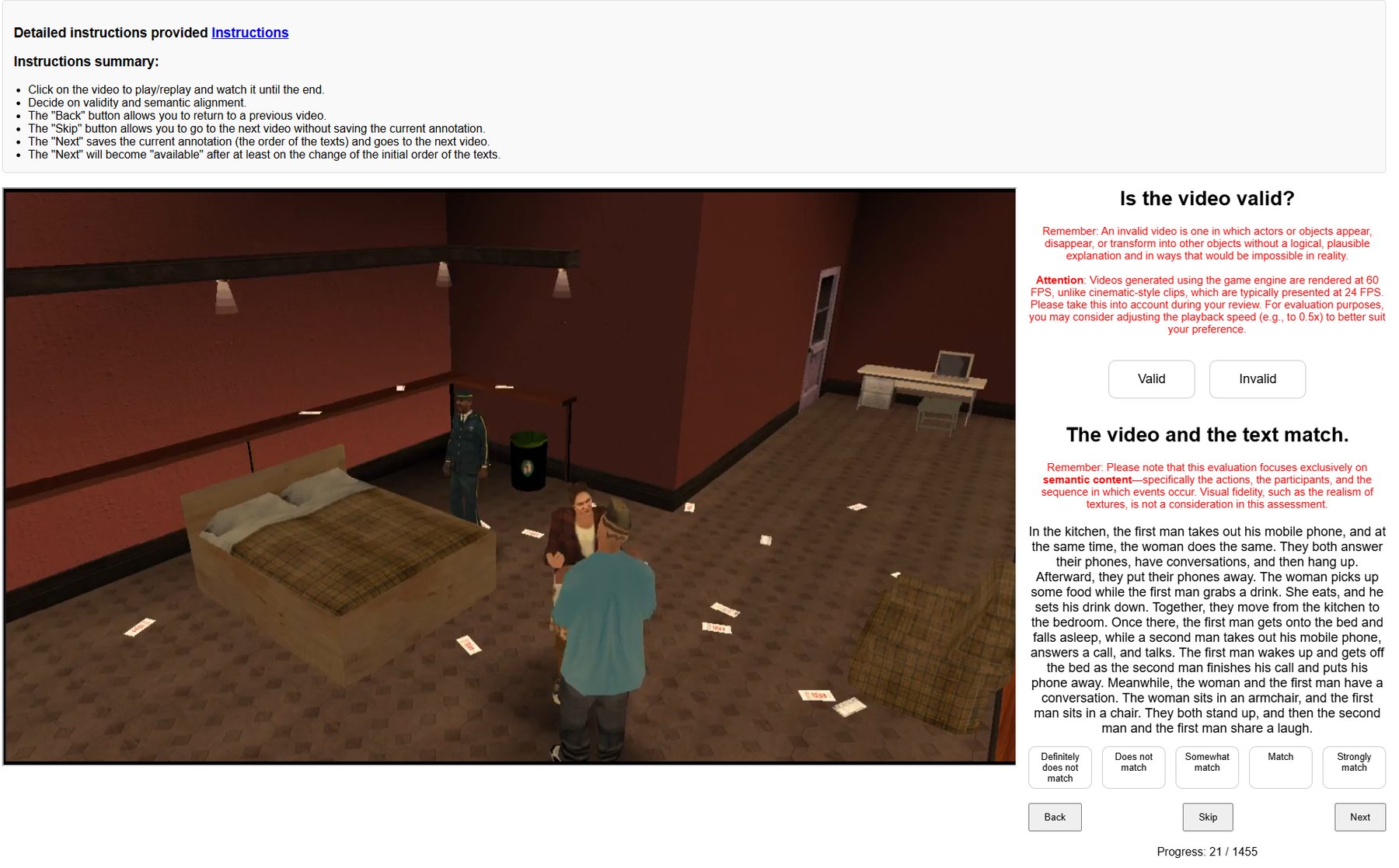}
   \caption{Snapshot of the validity and semantic match annotation interface. The video is placed centrally, with instructions (and link towards the full guidelines) at the top. Questions and the text are presented in the right part, with small instructions reminders for annotators.}
   \label{fig:example}
 \end{figure}

Both human studies (validity and semantic alignment; text description) use the same core web interface. The only difference between them is related to what is asked from the user: either answer two questions (is the video valid, how well does the text and video align semantically) or provide a text description. In Figure~\ref{fig:example} we provide a snapshot of the first interface, while the second version of the interface changes the questions with a text bot in which the user than describe the text. We present the videos in a randomized order while maintaining three-by-three groupings (i.e., GEST, VEO, and WAN) to ensure a fair comparison. Annotators are explicitly instructed to evaluate complete groups. They are also allowed and encouraged to revisit previously viewed videos when necessary to calibrate their assessment of the video--text semantic alignment.

\subsection{Instruction for Validity and Semantic Alignment}

The general guidelines used for the validity and semantic alignment human study are presented below:

\begin{verbatim}

You will be presented with a diverse set of video clips together
with a question regarding validity and a textual description.
Your task is to carefully watch the video, decide on its validity
and read the associated text (which is given an input to different
video generation models).

Validity
We define an invalid video as one in which actors or objects appear,
disappear, or transform into other entities without a logical,
plausible explanation, or in ways that would be physically impossible
in reality. In assessing validity, we do not consider semantic
consistency; for example, a sequence showing an actor sleeping while
simultaneously dancing would not be necessarily classified as invalid.

Semantic Alignment
In the second part, you will be presented with a text. Your task is
to evaluate the degree of semantic alignment between the text (input)
and the video (output), specifically assessing how accurately the
video reflects the requirements described in the text. The evaluation
should focus exclusively on semantic correspondence, including the
actors (number and gender) and the actions (presence and sequence).
Please disregard texture quality, video duration, and all other
visual attributes.

All the previously mentioned criteria are general guidelines and they
do not cover the whole spectrum of possible descriptions. As such, we
advise the users to use them as what they are, guidelines, and use
their best judgment when ranking the descriptions.
\end{verbatim}

\subsection{Instruction for Text Description}

The general guidelines used for the text description human study are presented below:

\begin{verbatim}

You will be presented with a diverse set of video clips and your task
is to carefully watch the provided video and write a detailed textual
description of the actions occurring in the video. Follow these
guidelines:

Describe all visible actions and movements performed by people or
objects. Write the description in a way that closely follows the
sequence of events in the video and mention actions even if they
seem small (e.g., turning, picking up an object). The description
should follow the chronological order of events. Do not summarize
the video broadly - describe what happens step-by-step as it unfolds.
If possible, mention the location or setting (e.g., kitchen, street,
office) and also reference relative positions when helpful (e.g.,
"on the table", "next to the door"). Please ignore visual quality
or style, do not comment on texture, visual realism, lighting and
artistic style and camera quality. The description should focus only
on what happens, not how it looks from a visual perspective. Describe
only what is visible, do not guess motivations, emotions or context
beyond what can be clearly observed. Finally, use a simple and
precise language and avoid vague summaries such as "things happens"
or "they interact".

Example:

Poor description: A realistic-looking video of a man doing something
in a kitchen.

Good description: A man stands in a kitchen next to a counter. He
picks up a knife from the counter and slices a tomato on a cutting
board. After cutting several slices, he places the knife down and
moves the tomato slices into a bowl.
\end{verbatim}

\section{Case Study: VLM Weak Points}
\label{sec:vlm_weak_points}

In Section 4.1 of the main paper we evaluated video validity with human annotators. We perform an additional analysis to assess the ability of existing Vision-Language Models (VLMs) to accurately identify invalid videos. Using the results of the human evaluation, we extract ground-truth validity labels and restrict the dataset to samples that received more than one annotation and achieved full annotator agreement. This filtering results in a total of 94 videos, with 45 valid and 49 invalid samples. In total we use 49 videos generated from GEST, 25 from VEO 3.1, and 20 from WAN 2.2. We run the same frontier models prompted with the same instruction used in the human study to check for video validity.

\begin{table}[!b]
  \caption{Frontier VLMs performance on video validity task}
  \label{tab:vlms_validity}
  \centering
  \scriptsize
  \begin{tabular}{@{}lcccc@{}}
    \toprule
    Model & Accuracy & Precision & Recall & F1 Score \\
    \midrule
    gpt-5.2-pro & 56.38 & \textbf{68.94} & 54.54 & 44.84\\
    gemini-3.1-pro & 56.38 & 59.76 & 57.44 & 54.09\\
    claude-opus-4.6 & 55.32 & 60.21 & 56.60 & 51.62\\
    \midrule
    Jury (maj.) &  \textbf{58.51} & 60.23 & \textbf{59.21} & \textbf{57.70}\\
  \bottomrule
  \end{tabular}
\end{table}

The performance of these models is reported in Table~\ref{tab:vlms_validity}. The results reveal a clear limitation of current frontier models on the validity task. The relatively low overall accuracy (around 56\%) and F1 score (around 50\%) suggest that this task captures phenomena that these models struggle to recognize. Specifically, events such as objects appearing or disappearing without cause, or actors morphing into entirely different individuals (e.g., from a brown-haired male to a blonde female), appear to be blind spots for current models. Moreover, this does not appear to disturb the models' interpretations: when asked to generate captions or descriptions for such videos, the models tend to ignore these inconsistencies and produce descriptions as if the content were coherent and natural. Finally, we note that constructing a majority-voting jury from the models leads to a slight improvement in performance, surpassing that of each individual model within the ensemble.

\section{VLMs as Jury Evaluation}
\label{sec:llm_jury}

In Section 4.2 we used a jury of frontier VLMs for evaluation (VLMs as Jury): each juror watches the video and scores the candidate descriptions. The descriptions were anonymized and provided in random order for each request to exclude any potential bias. The complete prompt is given below:

\begin{verbatim}
You will be presented with a video clip and {num_texts}
textual descriptions (labelled {labels}). Your task is
to watch the video, read all descriptions and give each
one a score between 0 and 100 based on the criteria below.
Scores must have NO ties - every description must receive
a different score so they can be unambiguously sorted
from best to worst.

## Scoring criteria (in order of importance)

1. **Richness**: the text should describe the actions
performed in the video in detail. We are not interested
in generally true statements that bring no meaningful
information (e.g. "a person exists"). For example if
in a video a person walks to a door and exits the room,
both "A person exits a room" and "A man walks towards the
door, opens it and then exits the room" are correct;
we prefer the latter as it contains more details.

2. **Factual correctness**: the text should accurately
and objectively describe what is happening in the video.
It should correctly describe the number of persons,
the order of actions, and not "hallucinate" details that
are not in the video (e.g. subjective affirmations such
as "preparing to go to bed" in a video where a person
just walks in a bedroom).

3. **Grammatical correctness**: the description should
be written in English and should not contain
grammatical errors.

A good textual description is factually correct,
rich and expressive enough that you could get a pretty
good idea of what happens in the video without
seeing it.

### Error severity (worst -> least severe)
- Missing persons (e.g. two actors in video but
description mentions only one)
- Missing actions (e.g. person talks on phone while
sitting but description only mentions sitting)
- Misidentified actions or objects (e.g. eating vs drinking,
sofa vs chair). Severity ranges from "very wrong" to "slightly wrong"
- punish serious errors while allowing small slip-ups where
objects or actions are similar.

These criteria are general guidelines. Use your best judgment.

## Descriptions

{descriptions}

## Required output format

Respond with EXACTLY this JSON and nothing else:
{output_schema}
\end{verbatim}

The reverse-matching study (main paper, Sec.~4.1) reused this same jury of models, prompted with the source description as anchor to select the closest of the three human-written descriptions.

\section{Additional Qualitative Captioning Cases}
\label{sec:qual_cases}

\begin{figure}[!t]
   \centering
   \includegraphics[width=\textwidth]{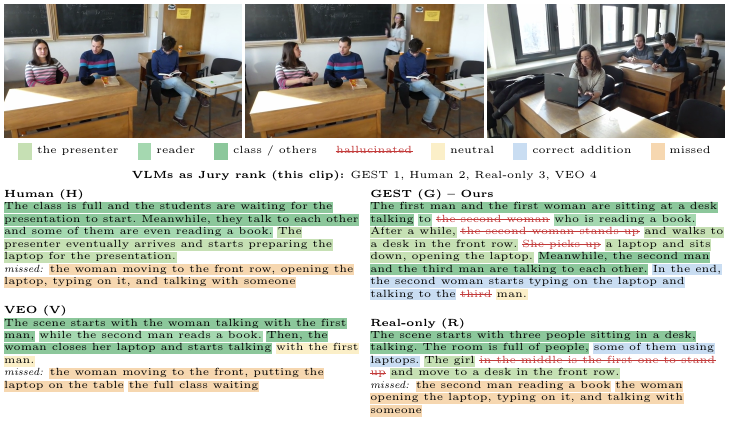}
   \caption{\textbf{Hand-verified qualitative case.} The jury ranks the G caption first, above the human reference. G recovers the full multi-actor sequence, including the presenter's move to the front, missed by H, despite identity and action slips (red, struck); V and R miss much of the story (orange).}
   \label{fig:qual_gexample2}
\end{figure}

Fig.~\ref{fig:qual_gexample2} shows the case discussed in the qualitative analysis of the main paper (Sec.~4.2). It and further positive and negative cases are presented in chapter~3 of the supplementary video, each shown as a clip with its full descriptions read against the playing video.

\section{Spatio-Temporal Probing Qualitative Examples}
\label{sec:probe_cases}

Chapter~4 of the supplementary video presents example clips for the spatio-temporal probing tasks of Sec.~4.3 of the main paper, including the previously untestable inter-entity relational tasks. Each clip is shown with the ground-truth label and the probe predictions of all six encoders.

\section{Frame Sampling Control}
\label{sec:frame_sampling_control}

\begin{table}[!t]
  \caption{Frame sampling control on Videos-to-Paragraphs (VideoLLaMA3 2B, mean$\pm$std over four seeds). GH: GEST with hybrid sampling (main paper); GU: GEST with uniform sampling; R: Real-only.}
  \label{tab:frame_sampling_control}
  \centering
  \scriptsize
  \setlength{\tabcolsep}{3pt}
  \begin{tabular}{@{}lccccccc@{}}
    \toprule
    & BLEU-4 & METEOR & CIDEr & ROUGE-L & SPICE & BERTScore & BLEURT \\
    \midrule
    GU & 12.7$\pm$.5 & 34.8$\pm$.2 & 27.1$\pm$7.2 & 36.4$\pm$.5 & 29.6$\pm$.3 & 90.7$\pm$.1 & 54.1$\pm$.4 \\
    GH & 13.9$\pm$.5 & 35.8$\pm$1.1 & 29.2$\pm$3.6 & 37.2$\pm$.5 & 30.4$\pm$1.2 & 90.8$\pm$.1 & 54.7$\pm$.4 \\
    R  & 12.2$\pm$.9 & 33.2$\pm$.9 & 25.5$\pm$5.9 & 35.3$\pm$.8 & 26.6$\pm$1.2 & 90.7$\pm$.1 & 53.3$\pm$.3 \\
    \bottomrule
  \end{tabular}
\end{table}

Our GEST conditions use event-aware hybrid sampling (GH), while we sample real and VEO videos uniformly at 1\,fps. To isolate the effect of frame selection, we train the GU control on GEST with the same uniform sampling (Table~\ref{tab:frame_sampling_control}). The ranking holds: GU wins 82\% of seed-metric comparisons against R and 61\% against V. Frame selection contributes, since GH leads GU on all seven means (79\%). The comparison disadvantages GEST. Real and VEO videos fit within the frame budget, and we sample them end to end at 1\,fps, whereas our scenarios average 29 events over 52\,s, and uniform sampling misses 20\% of these events entirely, while hybrid sampling covers every one. The event-to-frame mappings do not exist for the other conditions. Equalizing the strategies therefore requires discarding ours, which makes GU a lower bound.

\section{Extending the Engine}
\label{sec:extension_effort}

The world editor and the extension workflow documented in the technical report~\cite{cudlenco2026gestengineeventgraphssynthetic} extend the semantic breadth of the engine. In our experience, adding a new context takes about 2\,h, an object template 1\,h, and an action 30\,min.

\textbf{Coverage.} The action vocabulary is bounded, but each video contains more than the specified entities: the engine also logs world objects native to each context, beyond our mapping. Variability remains large. A scenario draws 2 to 6 actors from 312 skins, fills object slots from 124 templates across 20 supertemplates (see the technical report~\cite{cudlenco2026gestengineeventgraphssynthetic}), and runs under 23 weather conditions and 8 times of day, which exceeds $10^{26}$ combinations before any action is chosen.

\end{document}